\documentclass{article}

\usepackage{arxiv}

\usepackage[utf8]{inputenc} 
\usepackage[T1]{fontenc}    
\usepackage{hyperref}       
\usepackage{url}            
\usepackage{booktabs}       
\usepackage{amsfonts}       
\usepackage{nicefrac}       
\usepackage{microtype}      
\usepackage{lipsum}

\usepackage{times}
\usepackage{epsfig}
\usepackage{graphicx}
\usepackage{amsmath}
\usepackage{amssymb}
\usepackage{gensymb}

\title{Learning from Videos with Deep Convolutional LSTM Networks}

\author{
  Logan Courtney \\
  Department of Industrial and Enterprise Systems Engineering\\
  University of Illinois\\
  Urbana-Champaign, IL \\
  \texttt{courtne2@illinois.edu} \\
   \And
 Ramavarapu Sreenivas \\
  Department of Industrial and Enterprise Systems Engineering\\
  University of Illinois\\
  Urbana-Champaign, IL \\
  \texttt{rsree@illinois.edu} \\
}

\begin{document}
\maketitle

\begin{abstract}
This paper explores the use of convolution LSTMs to simultaneously learn spatial- and temporal-information in videos. A deep network of convolutional LSTMs allows the model to access the entire range of temporal information at all spatial scales of the data.  We describe our experiments involving convolution LSTMs for lipreading that demonstrate the model is capable of selectively choosing which spatiotemporal scales are most relevant for a particular dataset. The proposed deep architecture also holds promise in other applications where spatiotemporal features play a vital role without having to specifically cater the design of the network for the particular spatiotemporal features existent within the problem. For the Lip Reading in the Wild (LRW) dataset, our model slightly outperforms the previous state of the art (83.4\% vs. 83.0\%) and sets the new state of the art at 85.2\% when the model is pretrained on the Lip Reading Sentences (LRS2) dataset.
\end{abstract}


\section{Introduction}

Learning from video sequences requires models capable of handling both spatial and temporal information. Due to the advent of large image datasets such as ImageNet \cite{imagenet},  there has been significant progress in the development of convolutional-based architectures for learning spatial features \cite{alexnet}\cite{vgg16}\cite{resnet}\cite{inception}. It is not surprising that almost all methodologies for video sequences revolve around convolutional networks combined with additional temporal elements. 

The origin of video based learning begins primarily with the task of human action recognition. \cite{largescalevideoclassification} saw success by stacking frames together before passing them through a pretrained convolutional network as well as processing frames individually with some variant of temporal pooling applied to the network output. \cite{twostream} saw larger improvements by passing RGB images as well as optical flow images through pretrained networks and combining the output as described earlier. \cite{beyondshortsnippets} leveraged the success of recurrent neural networks by creating a network of stacked LSTMs \cite{LSTM} to process the output of a convolutional network. All of these methods for handling the temporal information used pretrained convolutional networks. The spatial and temporal features were, in some sense, handled separately.

\cite{spatiotemporal3d} first explored the use of 3D convolutions for processing spatial and temporal information together. That is, the network was capable of learning spatiotemporal features. Research has continued predictably along this path with large action recognition datasets such as the Kinetics dataset \cite{kinetics} and the Youtube-8M dataset \cite{youtube8m}. \cite{3dcnnretrace} replaced 2D convolutions in common image-based architectures such as ResNet with 3D convolutions. \cite{quovadis} was able to expand the filters from a pretrained 2D network to a 3D convolution network capturing the benefits of the extra training data from ImageNet. 

Lipreading is a technique for understanding speech using only the visual information of the speaker. Although there are clear applications such as speech transcription for cases where audio is not available, it is also a well-structured problem for looking at how deep networks learn spatiotemporal features. For a problem like action recognition, the current datasets have classes that can be identified from a single image alone (e.g. playing baseball versus swimming). In such instances, the temporal information is less important than the spatial information reducing the necessity of an architecture capable of handling spatiotemporal features. The current state of the art on the Kinetics dataset separates the spatial and temporal learning. \cite{revisitingofftheshelf} temporally post-processes learned features from a pretrained Inception-ResNet-v2 \cite{inceptionresnet} image model achieving higher performance than any of the 3D convolution architectures from \cite{3dcnnretrace}.

On the other hand, lipreading from a single frame within a sequence provides very little information about what is being said. It is a necessity to utilize the temporal context. Lipreading seems to be more reflective of the concept of a spatiotemporal problem.  This research explores the use of convolution LSTMs for lipreading. That is, 2D convolutions are used within the LSTM structure providing the network with the capacity to learn features at many combinations of spatial and temporal scales. The primary contributions of the paper are as follows.
\begin{itemize}
  \item This is the first very deep network built primarily with convolutional LSTMs and it achieves an improvement over the previous state of the art on the Lip Reading in the Wild (LRW) dataset \cite{LRW}
  \item Convolutional LSTMs see the same improvements as 2D convolutions see in image classification tasks when upgrading from architectures like VGG to ResNet. This is similar to what \cite{3dcnnretrace} demonstrated for 3D convolutions.
  \item The model utilizes 2D convolutions along with convolutional LSTMs demonstrating the ability to intermix temporally capable modules with spatial-only processing modules reducing computation and GPU memory usage
  \item A technique providing insight on what spatiotemporal scales are relevant to a particular problem and how this information can be used (Section \ref{section:sensitivityanalysis})
\end{itemize}

Section \ref{section:background} reviews related work in the literature along with some background information. Section \ref{section:proposedtechnique} describes our proposed methodology and model architectures. Section \ref{section:experiments} describes the experiments and implementation. Section \ref{section:results} compares our results with previous research and provides empirical evidence explaining the similar performance between the convolutional LSTM models and the current state of the art by analyzing the importance of particular spatiotemporal scales for lipreading.

\section{Background and Related Work}\label{section:background}

\subsection{Spatiotemporal Features and Receptive Field}
A network made up of convolutions takes an input image with a certain number of channels (typically three for RGB images) and has a height/width related to the spatial dimensions. Each layer within a convolutional network has an output which can be interpreted as an image with channels made up of a prescribed number of features (dependent on the network architecture) and a corresponding height/width related to the spatial dimensions. Spatial pooling and/or strided-convolution layers are used throughout the network to reduce the height/width of these output feature maps. This helps reduce computational costs, as there would be less convolution operations per layer; while increasing the visual receptive field, as each pixel of an output feature map is calculated based on a larger portion of the original input image.  Additionally, it allows for deeper (i.e. more layers) as well as wider (i.e. more channels) networks which have been shown to increase the network's capability to learn complex visual tasks.

However, with each spatial pooling layer and/or strided convolution layer, a certain degree of spatial information is inevitably lost due to a reduction in the height/width of the feature map output.  In addition, it potentially impedes the network's ability to learn functions capable of discriminating high resolution visual information. The choice of when to apply spatial dimension reduction techniques within the network remains an ``art-form'' in the hands of the network designer.  An application such as classifying large objects taking up the entire input image can utilize more spatial reduction layers to increase the visual receptive field, without discarding much relevant information. On the other hand, using many spatial reduction layers (especially too early within the network) may prevent the network from being able to detect the precise location of objects taking up a small portion of the input image. The receptive field of convolutional networks has been well studied in the past \cite{receptivefield}.

For 1D sequence problems, such as those seen in natural language processing (NLP), there are typically two techniques used to capture sequence information: convolutions and recurrent neural networks. Similar to the visual receptive field of 2D convolutions applied to images, a series of 1D convolutions on 1D data makes each subsequent layer see more context from the original input. That is, the temporal receptive field increases as more convolutions are applied. For a task like document classification, this gradual exposure to the sequence information with deep convolutional networks works well in practice \cite{deepcnntext}. On the other hand, each layer of a network made up of LSTMs has access to the full sequence. That is, the temporal receptive field when calculating the output at a particular timestep includes all previous inputs. This technique works well when training language models for predicting the next word in the sequence \cite{languagemodel}.

Dealing with sequences of images creates an additional challenge due to the temporal information appearing at multiple spatial scales. Our work is meant to motivate a principled approach for applications that depend on detecting these spatiotemporal features. It blends methods that (a) involve just images, and (b) involve just time series data.  So far, methods for applications like action recognition or lipreading have utilized a straightforward recipe of using 3D convolutions together with time-tested techniques from the two individual fields. As shown in \cite{spatiotemporal3d} for action recognition, utilizing a 3D convolution ResNet50 over a 2D convolution ResNet50 model with an LSTM shows an increase from 68.0\% to 72.2\%. Utilizing a recurrent neural network after the final layer of a 2D convolution network may be too late to capture the relevant spatiotemporal features. In \cite{3dresnetlstm} for lipreading, including a 3D convolution at the front of the network sees an increase of 5.0\% compared to the model without any temporal processing during the spatial processing. At first glance, it may seem that any architecture capable of processing spatial and temporal features together is all that is needed.

However, heavily related events may be separated by temporal gaps like watching a TV show with commercials or planning long-term strategy in board games. There is no inherent reason for treating the temporal dimension as simply an extra spatial dimension as is done with 3D convolutions.

Additionally, there are performance discrepancies with the use of 3D convolutions. In action recognition, \cite{spatiotemporal3d} saw success by using a model with the temporal receptive field increasing at the same rate as the spatial receptive field. In lipreading, \cite{LRW} saw this same method produce significantly worse results (over 10.0\% difference) when compared to processing all of the temporal information at a single spatial scale. Between the two higher performing models combining the temporal information at a single spatial scale, there was an additional 4.0\% improvement when the temporal processing was delayed until a later spatial scale. These discrepancies suggest different applications contain different spatiotemporal features and performance is dictated by which spatiotemporal features the model is designed to handle.

If the spatial resolution of the input doubled, would the model need to adapt the location of the temporal processing? If the temporal resolution of the input doubled, would a larger kernel size or more 3D convolutions at a particular spatial scale be necessary? There are still many unanswered questions and concerns with 3D convolutions. As much as deep learning is about creating networks capable of generalizing features relevant to a particular problem, it would be advantageous to create architectures capable of generalizing well across problems. This is the main focus of our work. The architecture presented in this paper and the empirical results suggest convolutional LSTMs hold promise in learning whichever spatiotemporal features are relevant to the dataset without having to specifically cater the design of the network for a particular set of spatiotemporal features.

\subsection{Lipreading in the Wild (LRW) Dataset}

The Lip Reading in the Wild (LRW) \cite{LRW} dataset consists of 500,000 videos with 29 frames each taken from BBC TV broadcasts. There are 500 target words (1,000 videos for each word) with 50 videos of each word for the validation set and 50 videos of each word for the test set. Due to the clips being slightly over 1 second long, very often there are context words surrounding the target word. Although there are ambiguous classes within the dataset (e.g. ``weather'' and ``whether''), it is sometimes possible to distinguish between these based on the small amount of surrounding context. The videos are centered on the speaker with the speaker facing the camera.

\subsection{Lipreading Sentences in the Wild (LRS2) Dataset}
The LRW dataset has a constrained vocabulary with fixed input size making it a well structured sequence classification problem. Lip reading becomes more difficult with an unconstrained vocabulary and variable length sequences. The Lip Reading Sentences (LRS) \cite{LRS} dataset contains ~118,000 sentences with a total of 17,428 unique words. The training set contains 16,501 unique words with the test set containing 6,882 words. There are words in the test set not seen in the training set. Additionally, the face direction is no longer constrained to facing the camera. There are moments of off-angle views up to 90\degree (side profile). The original LRS dataset is unreleased to the public due to licensing issues and a slightly smaller LRS2 dataset \cite{deepaudiovisual} is used in its place.

\subsection{Related Work}
\begin{figure}[t]
\begin{center}
    \includegraphics[width=0.2\linewidth]{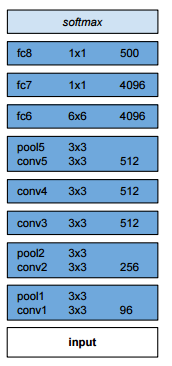}
\end{center}
   \caption{Modified versions of this VGG-M architecture were explored for lipreading in \cite{LRW}}
\label{fig:VGG-M}
\end{figure}

\begin{figure}[t]
\begin{center}
    \includegraphics[width=0.4\linewidth]{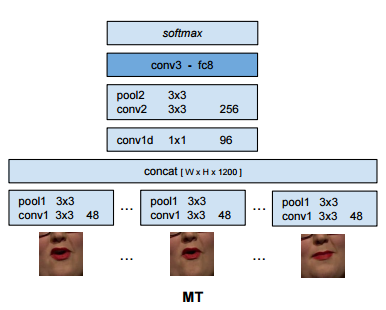}
\end{center}
   \caption{Original model setting the benchmark with the release of the LRW dataset at 61.1\% accuracy. A multitower (MT) structure is apparent due to processing each frame individually with a 2D convolution before concatenating the features for temporal processing.}
\label{fig:MT_2}
\end{figure}

The original LRW \cite{LRW} paper discussed the creation of the dataset and tested multiple variations of the VGG-M model (seen in Figure \ref{fig:VGG-M}). Their highest performing model (seen in Figure \ref{fig:MT_2}) processed each frame of the video with a 2D convolution before concatenating the outputs allowing the remainder of the spatial processing to have access to the full temporal receptive field. This outperformed the networks utilizing 3D convolutions which gradually increased the spatial and temporal receptive fields as the input passed deeper into the network.

The original LRS \cite{LRS} paper used a similar VGG-M based model except it limited visual the front-end of the network to a temporal receptive field of five. This network acted as a sliding window across the sequence providing a separate output at each timestep. Long-term temporal information was managed by using an LSTM encoder-decoder network with an attention mechanism \cite{attention} to spell words one character at a time. After training, the model was fine-tuned on the LRW dataset and it achieved an accuracy of 76.2\%. The 15.1\% improvement over the model in Figure \ref{fig:MT_2} is due to some combination of the different temporal receptive field and the extra training data from the LRS dataset.

\begin{figure}[t]
\begin{center}
    \includegraphics[width=0.4\linewidth]{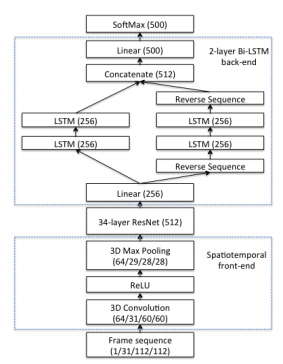}
\end{center}
   \caption{Current state of the art from \cite{3dresnetlstm} achieving 83.0\% accuracy on the LRW dataset. The spatiotemporal front-end uses a 3D convolution before being passed into a 34-layer ResNet. The output is then fed into a bidirectional LSTM for classification.}
\label{fig:3d_res_lstm}
\end{figure}

\cite{3dresnetlstm} replaced the VGG-M portion of the network with the deeper 34-layer ResNet \cite{resnet}. This network is separated into three parts: a spatiotemporal front-end made up of a 3D convolution with a temporal kernel of size five, ResNet34 for processing the remaining spatial information, and a bidirectional LSTM \cite{bidirectionallstm} for handling long-term temporal features. It is the current state of the art for the LRW dataset with 83.0\% accuracy. The model is shown in Figure \ref{fig:3d_res_lstm}.

\cite{deeplipreading} uses the same 3D convolution with ResNet34 architecture described above as the spatiotemporal front-end while testing three different back-end models for sequence transcription on the LRS2 dataset. The self-attention based Transformer \cite{transformer} outperformed both a fully convolutional back-end and a bidirectional LSTM back-end. The LR3-TED dataset \cite{lrs3ted} has recently been released and is roughly twice as large as the LRS2 dataset. \cite{deepaudiovisual} is a continuation of the work in \cite{deeplipreading} on this larger dataset and compares the performance of the Transformer back-end when trained as a seq2seq \cite{seq2seq} model versus training with the CTC loss \cite{CTC}. The seq2seq Transformer sets the benchmark at 48.3\% word error rate (WER) for LRS2 and 58.9\% WER for LRS3-TED.

It is important to note the above lipreading sentences research all use the same 3D+ResNet34 spatiotemporal front-end model (current state of the art on the LRW dataset) and they only compare back-end performance. The front-end model is pretrained on short sequences from the LRW and LRS datasets using the technique from \cite{lipreadinginprofile}. The back-end is then trained on extracted frozen features from the spatiotemporal front-end. The results illustrate the importance of surrounding context and language modeling when lipreading sentences. \cite{deepaudiovisual} shows over a 12\% reduction in WER when testing on phrases containing more than three words. The Transformer model contains over three times as many parameters (65 million) as the spatiotemporal front-end (21 million). The focus remained primarily on the long-term temporal context and no improved results were reported for the LRW dataset.

This is mentioned to emphasize the convolutional LSTM models explored here can be used in conjunction with these techniques by swapping out the spatiotemporal front-end. The back-end performance when transcribing longer sentences is directly related to the ability of the spatiotemporal front-end to extract high-quality features.

\section{Proposed Technique}\label{section:proposedtechnique}

\subsection{Convolution LSTM}

As opposed to a basic convolution layer, the convolutional LSTM contains an internal cell state $c_t$ (cf. equation \ref{equation1} below) and calculates a hidden state $h_t$ utilized as the output for subsequent layers as well as for state-to-state transitions. While processing a video sequence, $c_t$ and $h_t$ can be viewed as images of appropriate size maintained by the network with relevant information based on what it has seen in the past. Learnable filters $W_{\bullet}$ with bias terms $b_{\bullet}$ are used to handle a new input $x_t$ along with the past information being used by learnable filters $U_{\bullet}$.

\begin{equation}\label{equation1}
\begin{split}
f_t &= \sigma(W_f*x_t + U_f*h_{t-1} + b_f) \\
i_t &= \sigma(W_i*x_t + U_i*h_{t-1} + b_i) \\
o_t &= \sigma(W_o*x_t + U_o*h_{t-1} + b_o) \\
c_t &= f_t \circ c_{t-1} + i_t \circ tanh(W_c*x_t + U_c*h_{t-1} + b_c) \\
h_t &= o_t \circ tanh(c_t)
\end{split}
\end{equation}

Convolutional LSTMs have been used in the past in rather limited ways. \cite{convlstmweather} first demonstrated success with predicting future precipitation maps. A shallow model with two layers was applied to the input at a single spatial scale. \cite{convlstmgesture} explored learning spatiotemporal features related to gesture recognition using mainly 3D convolutions followed by a few layers of convolutional LSTMs for longer context. \cite{convlstmviolence} used a bidirectional version on the output of a 2D VGG13 \cite{vgg16} network to detect violence in videos. The majority of the spatial processing occurs before the convolutional LSTM layers are applied. The datasets for these applications are relatively small which may explain convolutional LSTMs limited use due to their tendency to overfit \cite{lstmoverfit}. However, recent large-scale datasets are providing an opportunity to explore their use in new ways.

There is a fundamental difference with the temporal receptive field when using convolutional LSTMs for spatiotemporal features as opposed to 3D convolutions. Successive 3D convolutions inherently limit long-term temporal information from being processed until deep in the network after the spatial dimension has been reduced. On the other hand, a deep network of convolutional LSTMs will have access to the full sequence at all of its spatial scales. The models here incorporate convolutional LSTMs throughout the entirety of the network and they demonstrate the existence of spatiotemporal features at multiple scales.

\subsection{VGG-M with Convolutional LSTMs}

\begin{figure}[t]
\begin{center}
    \includegraphics[width=0.4\linewidth]{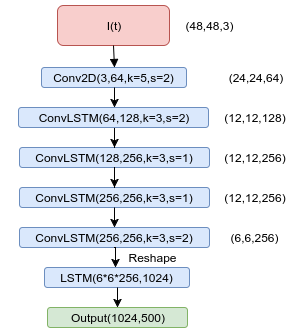}
\end{center}
   \caption{Baseline convolutional LSTM architecture modeled after the VGG-M architecture used in \cite{LRW}. }
\label{fig:baseconvlstm}
\end{figure}

As a baseline for comparison, the first convolutional LSTM model is based on the VGG-M architecture seen in Figure \ref{fig:VGG-M}. The number of output channels for the feature maps has been slightly reduced to compensate for the larger number of weights in a convolutional LSTM layer. The first layer is a regular 2D convolution which matches the highest performing multitower (MT) model from \cite{LRW} as shown in Figure \ref{fig:MT_2}. Batch normalization \cite{batchnorm} and a rectified linear unit (ReLU) were applied after every layer. The model is shown in Figure \ref{fig:baseconvlstm}.

\subsection{ResNet with Convolutional LSTMs}\label{section:resnet_conv_lstms}

Residual nets have been shown to be easier to optimize as well being able to create much deeper networks resulting in a significant boost in performance on image classification tasks \cite{resnet}. Certain convolution layers are replaced with convolutional LSTM layers extending the ResNet architecture to learn spatiotemporal features.

\begin{figure}[t]
\begin{center}
    \includegraphics[width=0.4\linewidth]{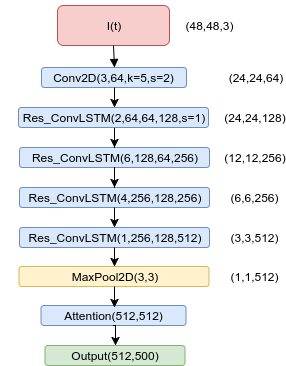}
\end{center}
   \caption{A residual based network incorporating convolutional LSTM layers. Each \textbf{Res\_ConvLSTM} layer is made up of multiple sub-blocks. Sublock A (seen in Figure \ref{fig:BlockA}) is when spatial downsampling is needed. Sublock B (seen in Figure \ref{fig:BlockB}) is used for all other cases.  The output dimensions are shown in the margins.}
\label{fig:resnet_convLSTM}
\end{figure}

\begin{figure}[t]
\begin{center}
    \includegraphics[width=0.4\linewidth]{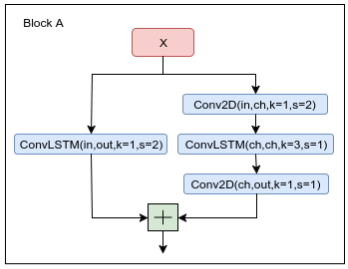}
\end{center}
   \caption{Block type A contains two convolutional LSTM layers with one of them in the path of the skip connection. This block has a stride of $2$ reducing the spatial dimensions by half.}
\label{fig:BlockA}
\end{figure}

\begin{figure}[t]
\begin{center}
    \includegraphics[width=0.4\linewidth]{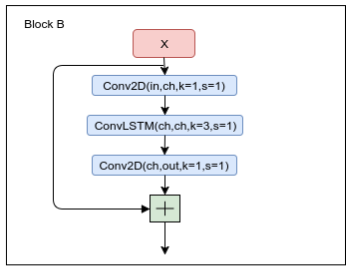}
\end{center}
   \caption{The second type of residual block contains one ConvLSTM layer with an open skip connection.}
\label{fig:BlockB}
\end{figure}

Figure \ref{fig:resnet_convLSTM} shows the model architecture. Batch normalization and ReLU layers are not shown but were used after every layer. The four parameters of a \textbf{Res\_ConvLSTM} module represent the number of sub-blocks, the number of input channels, the number of intermediate channels, and the number of output channels. For any \textbf{Res\_ConvLSTM} layer, the first sub-block is always of type A (shown in Figure \ref{fig:BlockA}) with the remaining sub-blocks of type B (shown in Figure \ref{fig:BlockB}). This is similar to the technique used in \cite{3dcnnretrace} to replace the 2D convolutions in ResNet with 3D convolutions.

There are two variants. The first is used with an LSTM layer when training on the LRW dataset in place of the attention mechanism seen in Figure \ref{fig:resnet_convLSTM}. The attention network \cite{attention} is used when training on the LRS2 dataset. The convolutional LSTM based ResNet model has 14.5 million parameters. The bidirectional version has 29 million parameters which is roughly equivalent to the 23.5 million parameters in the current top performing model shown in Figure \ref{fig:3d_res_lstm}. There are 48 total layers: 14 3x3 convolutional LSTMs, 4 1x1 convolutional LSTMs, 28 1x1 2D convolutions, and 1 LSTM before the final output layer.

\subsection{Connectionist Temporal Classification (CTC)}
Due to the unconstrained vocabulary of the LRS2 dataset, the network is modified to output a character as well as the one additional blank token necessary for using the CTC loss \cite{CTC}. With this setup, the model learns to spell words one character at a time.

\section{Experiments}\label{section:experiments}

\subsection{Training Convolutional LSTMs on LRW} \label{section:lrw_train}
A random crop around the mouth between 48 and 68 pixels is attained for each frame. This is resized to a 48x48 input. Note this crop is less than half the size of the 112x112 crop used in \cite{LRW}\cite{LRS}\cite{3dresnetlstm}\cite{deepaudiovisual}\cite{deeplipreading}. This reduction in input size greatly reduces the GPU memory overhead and reduces the computation to a manageable amount while still maintaining enough detail to achieve high performance. During training, the input frames are randomly flipped, randomly rotated +/- 10 degrees, and have the brightness adjusted randomly by +/- 10\%. The model is also trained on random subsequences of length 24 as opposed to the full 29 frames. The word boundaries are unknown but this worked well to reduce overfitting. Dropout for recurrent neural networks \cite{lockeddropout} was used with $p=0.5$ before the final fully connected LSTM layer as an additional form of regularization. This means the same dropout mask was used for the entire sequence as opposed to the mask being generated independently at each timestep.

The internal cell states $c_t$ and the hidden states $h_t$ are reset to $0$ for each layer before processing a new sequence. The model produces an output after every timestep with the loss function applied to each one. Although the word can almost certainly not be correctly identified early in the sequence, this allows the use of truncated backpropagation. The loss is calculated and backpropagation is performed every 8 timesteps. This does not mean temporal features longer than 8 frames cannot be learned. The internal cell states and hidden states still carry old information even after the gradients are calculated. Backpropagation is performed three times per sequence with the parameter update happening only once at the end of the sequence. The GPU memory use scales linearly with the length of truncated backpropagation meaning the batch size can be increased for faster training.

Both the VGG-M based convolutional LSTM model and the ResNet based convolutional LSTM model took approximately three weeks to train with PyTorch \cite{pytorch} on a NVIDIA Titan X GPU. The network was trained using Adam \cite{adam} with stochastic gradient descent. An initial learning rate of $1e^{-4}$ was used and reduced whenever performance on the validation set stopped progressing. Near the end of training, the sequence length was stepped up from 24 frames to the full 29 frames in order to allow the model to take more advantage of the context. This increased performance by 2\%-3\%.

A bidirectional version of the ResNet based model was trained as well. However, they were not trained simultaneously and the reverse direction network was initialized with the already trained forward direction network. The network rapidly learned the reverse sequences and increased the accuracy by 1.9\% when tested in conjunction with the forward direction network.

\subsection{Pretraining on LRS2}

As mentioned in section \ref{section:resnet_conv_lstms}, the output of the ResNet based convolutional LSTM was modified to predict characters in order to leverage additional training with the LRS2 dataset. The CTC loss function is calculated with the full sequence output without truncated back propagation. The increase in GPU memory use from this limits the model to a batch size of 20 when the sequence length is 24. Gradient clipping \cite{gradientclipping} was set to $5.0$. This was necessary due to the occasionally large gradients. 

Curriculum learning was used to speed up training and has been shown to improve results \cite{curriculumlearning}\cite{LRS}. The pretraining dataset contains labeled word boundaries. Training began on subsequences of length 8. This was gradually increased until a sequence length of 24. After training for approximately two weeks, learning slowed significantly. The model was then fine-tuned on the LRW dataset. This took approximately two more weeks. The trained model was used to initialize and train a reverse direction network just as was done in section \ref{section:lrw_train}.

\section{Results}\label{section:results}

During inference, three crops of size 48x48, 56x56, and 64x64 with their flipped counterparts (six transformations total) are passed through the network with the outputs averaged for the final prediction. The results are summarized in Table \ref{table:results}. The bottom four entries are the convolutional LSTM models trained here.

\begin{table*}[]
\centering
\caption{Results comparing the performance on the LRW dataset. The bidirectional ResNet with convolutional LSTMs achieves state of the art results with and without LRS2 pretraining.}
\label{table:results}
\begin{tabular}{|l|c|l|l|l|l}
\cline{1-5}
                                      & Pretrained on LRS2?& Top-1   & Top-5   & Top-10  &  \\ \cline{1-5}
VGG-M Multitower \cite{LRW}           & no                 & 61.1\%  & -       & 90.4\%  &  \\ \cline{1-5}
VGG-M+LSTM \cite{LRS}                 & yes                & 76.2\%  & -       & -       &  \\ \cline{1-5}
                                      &                    &         &         &         &  \\ \cline{1-5}
2D+ResNet34+Conv \cite{3dresnetlstm}  & no                 & 69.6\%  & 90.4\%  & 94.8\%  &  \\ \cline{1-5}
3D+ResNet34+Conv \cite{3dresnetlstm}  & no                 & 74.6\%  & 93.4\%  & 96.5\%  &  \\ \cline{1-5}
3D+ResNet34+BiLSTM \cite{3dresnetlstm}& no                 & 83.0\%  & 96.3\%  & 98.3\%  &  \\ \cline{1-5}
                                      &                    &         &         &         &  \\ \cline{1-5}
VGG-M ConvLSTM                        & no                 & 73.1\%  & 92.5\%  & 96.7\%  &  \\ \cline{1-5}
ResNet ConvLSTM                       & no                 & 81.5\%  & 96.1\%  & 98.2\%  &  \\ \cline{1-5}
ResNet BiConvLSTM                     & no                 & 83.4\%  & 96.8\%  & 98.5\%  &  \\ \cline{1-5}
ResNet BiConvLSTM                     & yes                & 85.2\%  & 97.4\%  & 98.9\%  &  \\ \cline{1-5}
\end{tabular}
\end{table*}

The VGG-M based convolutional LSTM model achieved 73.1\%, higher than all of the spatiotemporal VGG-M models from \cite{LRW}. This result is less pronounced when comparing the residual based architectures.

The first two ResNet34 models in table \ref{table:results} are modifications of the model shown in Figure \ref{fig:3d_res_lstm}. 2D+ResNet34+Conv did not use a spatiotemporal front-end and used 1D convolutions over the temporal dimension for the back-end to achieve 69.6\% accuracy. The front-end was replaced with a 3D convolution with an increase of 5.0\% in performance (74.6\%) and the final model used a bidirectional LSTM back-end for another 8.4\% increase in performance (83.0\%). The second model closely resembles the VGG-M multitower model demonstrating the ResNet34 architecture is most likely responsible for much of the 13.5\% difference.

The bidirectional ResNet with convolutional LSTMs slightly outperforms the previous state of the art (83.4\% vs. 83.0\%). The same model sets the new state of the art with 85.2\% accuracy after pretraining on the LRS2 dataset.

\begin{table}[]
\centering
\caption{Top 5 best and worst performing classes. Eighteen of the words were perfectly classified.}
\label{table:bestworst}
\begin{tabular}{|l|l|l|l|}
\hline
WESTMINSTER     & 100\% & THINK & 50\% \\ \hline
INVESTMENT     & 100\% & UNDER & 52\% \\ \hline
CAMERON     & 100\% & UNTIL & 52\% \\ \hline
CAMPAIGN    & 100\% & THERE & 54\% \\ \hline
THOUSANDS   & 100\% & COULD & 56\% \\ \hline
\end{tabular}
\end{table}

Words with the highest and lowest classification rates are shown in Table \ref{table:bestworst}. Words with three or four syllables had an accuracy of 90.3\% while words with one syllable had an accuracy of 79.4\%. The network may not need to recognize the entirety of the word with multiple syllable words providing more opportunities for the classifier to correctly discriminate. The most confused words shown in Table \ref{table:confusion} all share features with less opportunity for discrimination. Due to the unknown word boundaries, it may be difficult for the classifier to correctly disentangle features belonging to the context words versus the target word making misclassifications of single syllable words and these similar words more likely. However, the context can be helpful. This explains how homophones like ``weather'' and ``whether'' have classification rates of 82\% and 92\%.

\begin{table}[]
\centering
\caption{Most confused word pairs.}
\label{table:confusion}
\begin{tabular}{|l|l|l|}
\hline
WORST    & WORDS    & 24\% \\ \hline
WANTS    & WANTED   & 20\% \\ \hline
PRESS    & PRICE    & 18\% \\ \hline
SPENT    & SPEND    & 18\% \\ \hline
THINK    & THING    & 18\% \\ \hline
\end{tabular}
\end{table}

\subsection{Spatiotemporal Features Sensitivity Analysis}\label{section:sensitivityanalysis}
The internal cell states and the hidden states are reset to 0 before processing a new sequence. Therefore, the outputs from the convolutional LSTM layers for the first frame have no prior information. The relative importance of temporal information at various spatial resolutions can be explored by artificially resetting the internal states during processing to cause these layers to respond as if it has no context.

\begin{figure}[t]
\begin{center}
    \includegraphics[width=0.6\linewidth]{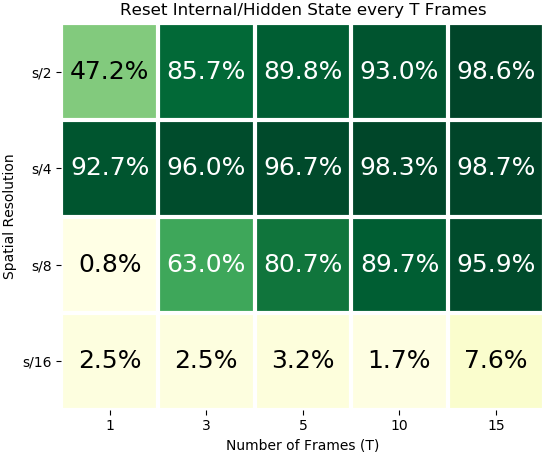}
\end{center}
   \caption{Relative drop in performance when convolutional LSTM internal state is reset every T frames at particular spatial scales.}
\label{fig:reset}
\end{figure}

The performance on a random subset of 5000 video sequences from the validation set is used as the metric for determining the importance of a particular spatiotemporal scale. Every T frames for $T=\{1,3,5,10,15\}$, the convolutional LSTMs at a single spatial scale are reset to 0 with all other layers operating normally. The relative drop in performance as opposed to the full network operating normally is shown in Figure \ref{fig:reset}. The final layer at $s/16$ shows poor performance anytime the temporal information is not used. This makes sense considering this where the network would use the long term information from the whole sequence right before classification. The $s/8$ scale shows poor performance when the internal state is reset every frame, moderately decreased performance when reset every 3 to 10 frames, and almost back to normal when it's reset every 15 frames. At this spatial scale, the model seems to learn sequences approximately 10 frames long. The $s/2$ scale drops significantly only when the state is reset every frame. This is at the highest spatial resolution and supports the intuition that the short term dynamics are important for lipreading. The $s/4$ scale seems to always perform well. The hidden to hidden operations provide virtually no additional information.

These empirical results support the architecture design of 3D+ResNet+BiLSTM and potentially show why the ResNet based convolutional LSTM performs so similarly. The bidirectional LSTM backend captures the long term dependencies while the 3D spatiotemporal front-end captures the short term dynamics. The convolutional LSTM model had access to the full temporal receptive field at all spatial scales but ultimately arrived at nearly the same architecture by simply not utilizing the hidden-to-hidden connections at $s/4$. There are 19 of the 48 layers at this scale.

Convolutional LSTMs may be beneficial from a design perspective allowing a quick approach for new applications when knowledge of the relevant spatiotemporal features is unknown ahead of time. It is easier to let the network figure out what is relevant from the data as opposed to designing the network a particular way. However, if features at a particular scale do not exist for the true underlying distribution, these additional connections simply provide more opportunity for the model to overfit. For this reason, it may still be beneficial to utilize a convolutional LSTM for discovering what features are important followed by specifically designing a network capable of handling only these specific features. For this particular task of lipreading, it seems the 3D+ResNet+BiLSTM model happens to naturally fit the features existing within the dataset.

\section{Conclusion}\label{section:conclusion}
A deep convolutional LSTM model is successfully trained and achieves state of the art performance for the Lip Reading in the Wild (LRW) dataset. The network is shown to have successfully learned relevant spatiotemporal features from the data without having to specifically cater the design of the network for the specific problem. Additionally, the benefits of using convolutional architectures like ResNet are apparent for Convolutional LSTMs just as they have been shown in the past for 2D and 3D convolutions.

There are still many problems facing convolutional LSTMs. They contain a large number of weights making their use computationally expensive as well as making them prone to overfitting. For many real applications, their ability to handle a larger array of spatiotemporal features may be unnecessary. However, as datasets become larger and more complex, we may need to rely on the capabilities of convolutional LSTMs. This is the first successful model of its kind and it shows promise for approaching new problems or being used to understand new datasets. 

Future work will explore utilizing convolutional LSTM models in conjunction with current state of the art techniques \cite{deeplipreading}\cite{deepaudiovisual} for lip reading sentences on new larger datasets like LRS3-TED \cite{lrs3ted}.

{\small
\bibliographystyle{unsrt}
\bibliography{template}
}

\end{document}